\documentclass[10pt,twocolumn,letterpaper]{article}

\usepackage{cvpr}              
\definecolor{cvprblue}{rgb}{0.21,0.49,0.74}
\usepackage[pagebackref,breaklinks,colorlinks,allcolors=cvprblue]{hyperref}
\usepackage{tabularx}

\def\paperID{37640} 
\def\confName{CVPR}
\def\confYear{2026}

\title{OpenTrack3D: Towards Accurate and Generalizable Open-Vocabulary 3D Instance Segmentation}

\author{
Zhishan Zhou\textsuperscript{1} \quad
Siyuan Wei\textsuperscript{1} \quad
Zengran Wang \textsuperscript{1} \quad
Chunjie Wang\textsuperscript{1} \quad
Xiaosheng Yan\textsuperscript{1} \quad
Xiao Liu\textsuperscript{1} \\
\textsuperscript{1}PICO, ByteDance, Beijing \\
\parbox{\linewidth}{\centering
{\tt\small \{zhouzhishan.2013, weisiyuan.buaa, wangzengran,\\
{\tt\small wangchunjie01, yanxiaosheng, liuxiao.ai\}@bytedance.com}
}
}
}

\begin{document}
\maketitle
\begin{abstract}
Generalizing open-vocabulary 3D instance segmentation (OV-3DIS) to diverse, unstructured, and mesh-free environments is crucial for robotics and AR/VR, yet remains a significant challenge. We attribute this to two key limitations of existing methods: (1) proposal generation relies on dataset-specific proposal networks or mesh-based superpoints, rendering them inapplicable in mesh-free scenarios and limiting generalization to novel scenes; and (2) the weak textual reasoning of CLIP-based classifiers, which struggle to recognize compositional and functional user queries.
To address these issues, we introduce OpenTrack3D, a generalizable and accurate framework. Unlike methods that rely on pre-generated proposals, OpenTrack3D employs a novel visual–spatial tracker to construct cross-view consistent object proposals online. Given an RGB-D stream, our pipeline first leverages a 2D open-vocabulary segmenter to generate masks, which are lifted to 3D point clouds using depth. Mask-guided instance features are then extracted using DINO feature maps, and our tracker fuses visual and spatial cues to maintain instance consistency. The core pipeline is entirely mesh-free, yet we also provide an optional superpoints refinement module to further enhance performance when scene mesh is available. Finally, we replace CLIP with a multi-modal large language model (MLLM), significantly enhancing compositional reasoning for complex user queries. Extensive experiments on diverse benchmarks, including ScanNet200, Replica, ScanNet++, and SceneFun3D, demonstrate state-of-the-art performance and strong generalization capabilities.

\end{abstract}

\begin{figure}[ht]
\centering
 \includegraphics[width=1.0\linewidth]{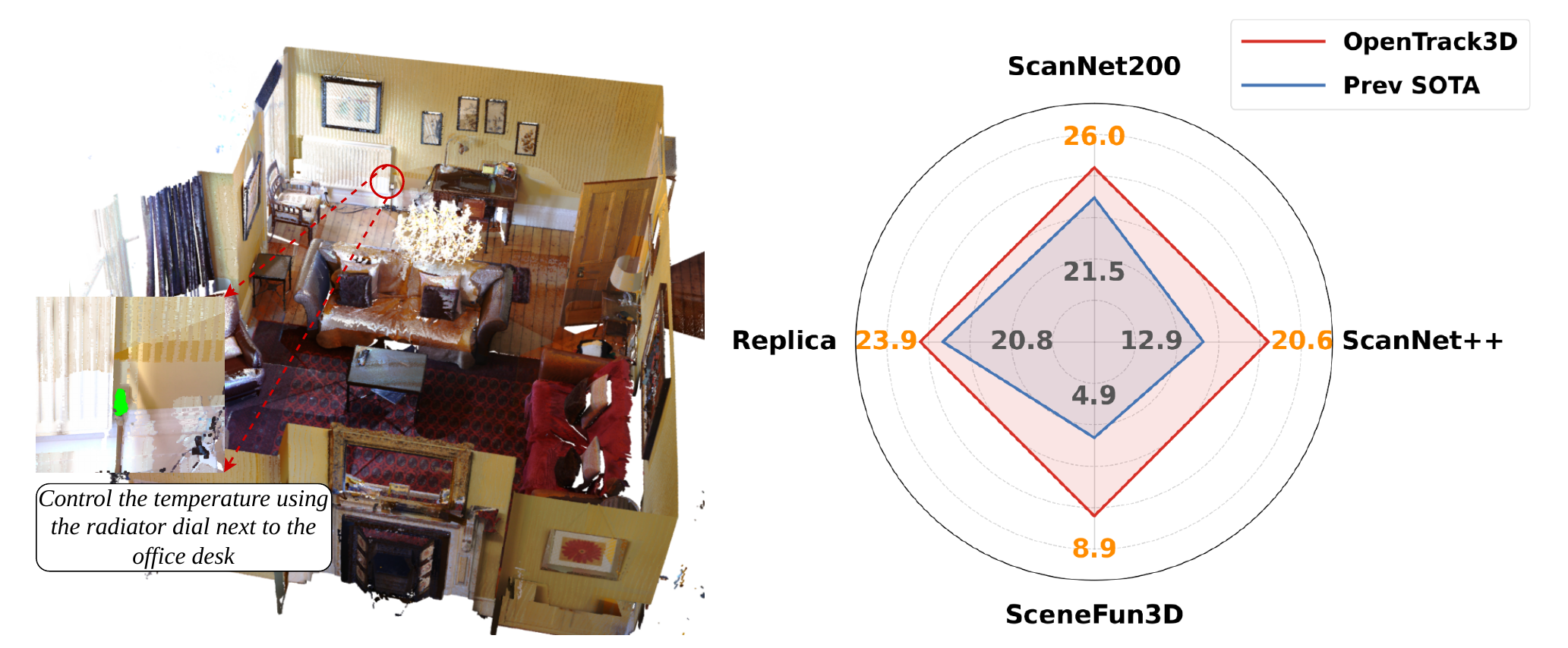} 
\caption{Qualitative and quantitative results of our method. 
\textbf{Left:} Visualization of our results on the \textbf{SceneFun3D} dataset. 
As shown, our method accurately localizes the target objects even under complex query. 
\textbf{Right:} Comparison with prior methods that do not rely on supervised 3D masks, where our approach consistently outperforms them across all benchmarks.}
\label{fig:radr_cmp}
\vspace{-0.2cm}
\end{figure}
\section{Introduction}
\label{sec:intro}

3D instance segmentation is fundamental to comprehensive scene understanding and underpins applications in VR/AR, robotics, and embodied AI. Unlike single-sensor 2D imaging, precise 3D perception typically depends on calibrated multi-view or RGB-D sensing, which makes large-scale, instance-level annotation costly. The resulting scarcity of labeled 3D data remains a central obstacle to advancing robust, deployable 3D perception systems.

Conventional 3D segmentation has therefore concentrated on a limited set of annotated categories, operating in a \textit{closed-vocabulary} regime. Methods span convolutional designs~\cite{vu2022softgroup,ngo2023isbnet,shin2024spherical,peng2024oa} and transformer-based architectures~\cite{schult2022mask3d,kolodiazhnyi2024oneformer3d}, and they process diverse 3D representations—from point clouds and voxels to neural fields~\cite{vora2021nesf,zhi2021place,chou2024gsnerf}. While these systems achieve strong results on benchmarks such as ScanNet200~\cite{rozenberszki2022language}, their practical use is constrained by the small label space and the limited availability of scene-level training data.

To relax vocabulary constraints and better leverage existing priors, recent work explores \textit{open-vocabulary} 3D instance segmentation (OV-3DIS). Many existing solutions are \textbf{training-based}, requiring task- or dataset-specific learning to transfer 2D foundation-model knowledge into 3D~\cite{guo2024sam,huang2024segment3d,peng2023openscene,ding2023pla,Ding_Yang_Xue_Zhang_Bai_Qi_2023,Yang_Ding_Wang_Qi_2024,chen2023clip2scene}. Typical pipelines rely on image-bridged supervision, SAM-derived pseudo labels, or distillation from powerful 2D models~\cite{radford2021learning,liu2024groundingdinomarryingdino,kirillov2023segment,ravi2024sam}. This introduces non-trivial training costs, sensitivity to pseudo-label quality, and limited scalability across domains and sensors.

The current generation of training-free OV-3DIS pipelines~\cite{huang2024openins3d,takmaz2023openmask3d,yan2024maskclustering,nguyen2025any3dis,boudjoghra2024open} relies on combining off-the-shelf 2D models with geometric cues, yet their architectures are constrained by inherent limitations. In proposal generation, existing methods fall into two restrictive paradigms: some utilize supervised proposals~\cite{takmaz2023openmask3d,huang2024openins3d,boudjoghra2024open,peng2023openscene}, which inherently restricts the model's generalization; while others~\cite{nguyen2025any3dis,yin2024sai3d,zhao2025sam2object,jung2025details,nguyen2024open3dis} depend on mesh-based superpoints, thereby compromising proposal granularity and becoming inapplicable in mesh-free datasets. Methods \cite{yan2024maskclustering,lu2023ovir} avoid the aforementioned issues but yield inferior performance. Concurrently, almost all prior works employ 2D CLIP-based~\cite{radford2021learning} features for classification. This reliance imposes a fundamental limitation, as CLIP lacks the compositional reasoning vital for complex 3D queries.

\begin{figure*}[t]
    \centering
    \includegraphics[width=0.9\linewidth]{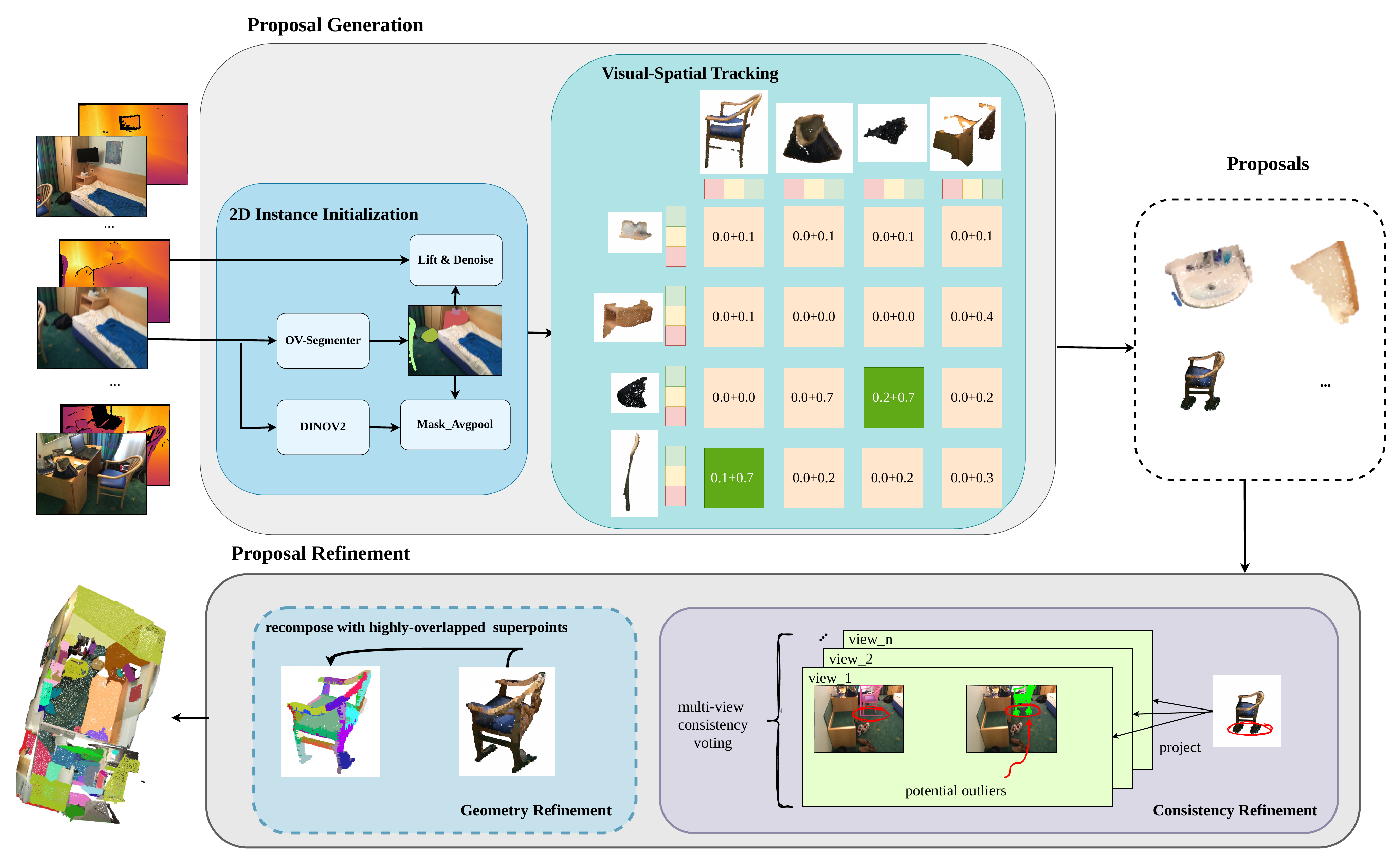}
    \caption{
        Overview of our \textbf{Proposal Generation} and \textbf{Proposal Refinement} stages. 
        \textbf{Proposal Generation} employs a 2D open-vocabulary segmenter to generate object masks, which are lifted to 3D point clouds using depth maps and subsequently denoised. 
        Mask-guided instance features are extracted from DINO feature maps, and our tracker fuses both visual and spatial cues to maintain instance consistency across frames. 
        \textbf{Proposal Refinement} further improves the 3D instances through \textit{Consistency Refinement} and \textit{Geometry Refinement}, while the merging process is omitted for clearer visualization.
    }
    \label{fig:ppl}
\end{figure*}

To address these issues, we introduce \textbf{OpenTrack3D}, a generalizable and accurate framework. 
Unlike methods reliant on supervised 3D proposals or mesh dependencies, OpenTrack3D employs a novel \textbf{visual–spatial tracker} to construct cross-view consistent object proposals online from an RGB-D stream.
~\cref{fig:ppl} provides an overview of the design:
in \textbf{Proposal Generation}, a 2D open-vocabulary detector~\cite{cheng2024yolo} and SAM~\cite{ravi2024sam} produce high-quality masks that are lifted to 3D. Our tracker then fuses visual (DINO features~\cite{oquab2023dinov2}) and spatial cues to maintain instance consistency across views, forming robust track-centric proposals. In \textbf{Proposal Refinement}, multi-view consistency filtering suppresses leakage and depth noise, and duplicate hypotheses are merged. Crucially, while the core pipeline is mesh-free, we also provide an \textbf{optional Geometry Refinement} to further enhance boundary fidelity when a scene mesh is available. 
In proposal classification, a small set of informative views per candidate is selected, and a multi-modal large language model (MLLM) is employed for open-vocabulary classification, enabling our system to understand and respond to complex user queries.
As shown in \cref{fig:radr_cmp}, our method achieves state-of-the-art results on multiple benchmarks, including ScanNet200~\cite{dai2017scannet}, ScanNet++~\cite{ScanNet++}, Replica~\cite{replica19arxiv}, and SceneFun3D~\cite{delitzas2024scenefun3d}, while demonstrating a strong capability to handle complex, fine-grained queries.

Our contributions are summarized as follows:
\begin{itemize}
    \item We propose \textbf{OpenTrack3D}, a \emph{training-free} framework for OV-3DIS. Its core is a novel \textbf{visual–spatial tracker} that generates proposals online by fusing visual and spatial cues, making the approach inherently mesh-free and generalizable.
    \item We introduce an \textbf{MLLM-based} recognition module that leverages top-$K$ informative views for each candidate and performs open-vocabulary classification under a flexible, task-specific label set, enabling robust understanding of complex user queries.
    \item Our framework achieves \textbf{state-of-the-art} results on four challenging and diverse benchmarks (ScanNet200, ScanNet++, Replica, and SceneFun3D), demonstrating strong generalization and high accuracy.
\end{itemize}
\section{Related Work}
\label{sec:related_work }

\paragraph{Open-Vocabulary 3D Instance Segmentation.} 
Despite remarkable progress in 2D open-vocabulary perception~\cite{liu2024groundingdinomarryingdino,cheng2024yolo}, analogous capabilities in 3D remain underexplored. Existing approaches can be broadly categorized into \textit{training-based} and \textit{training-free} paradigms. Training-based methods~\cite{guo2024sam,huang2024segment3d,peng2023openscene,ding2023pla,Ding_Yang_Xue_Zhang_Bai_Qi_2023,Yang_Ding_Wang_Qi_2024,chen2023clip2scene} typically transfer 2D model capabilities into the 3D domain through distillation or pseudo-labeling, but their applicability is constrained by the scarcity of large-scale 3D annotated data. In contrast, training-free methods mainly differ in how they generate 3D proposals: some rely on closed-vocabulary supervised networks such as~\cite{takmaz2023openmask3d,huang2024openins3d,boudjoghra2024open,peng2023openscene, zhou2025ov3d}, while others~\cite{yin2024sai3d,nguyen2024open3dis,nguyen2025any3dis,zhao2025sam2object,jung2025details} leverage superpoints extracted from global scene meshes. MaskClustering~\cite{yan2024maskclustering} and OVIR-3D~\cite{lu2023ovir} directly reconstruct point-level proposals from individual frames, though their performance remains inferior to the aforementioned counterparts. For open-vocabulary instance classification, most prior works follow OpenMask3D~\cite{takmaz2023openmask3d}, employing CLIP-based representation in Open3DIS~\cite{nguyen2024open3dis} or AlphaCLIP~\cite{sun2024alpha} in DetailMatters~\cite{jung2025details}. While OV3D-CG~\cite{zhou2025ov3d} also explores MLLMs, it relies on intermediate description generation, leading to a more complex pipeline. In contrast, our approach directly leverages an MLLM for instance-level classification, enabling stronger semantic understanding with a simpler framework.

\paragraph{Functionality and Affordance Segmentation.} 
The task of functionality and affordance segmentation was first introduced by SceneFun3D~\cite{delitzas2024scenefun3d}. Unlike prior datasets such as ScanNet200~\cite{dai2017scannet} and ScanNet++~\cite{ScanNet++}, it requires segmenting extremely small \textit{functional elements}, such as \textit{buttons}, \textit{handles}, and \textit{light switches}. Moreover, the provided textual descriptions are not directly tied to specific object names—for example, queries like ``flush the toilet'' require the model to possess strong language understanding and reasoning capabilities. SceneFun3D proposed a baseline that trains a specialized Mask3D network on its task-specific training set, and then adopts the CLIP-based classification strategy from OpenMask3D to determine whether each segmented object aligns with the given task description. Fun3DU~\cite{corsetti2025functionality} further introduces a large language model (LLM) to decompose the task into two parts: identifying the \textit{contextual object} (shown in pink) and the \textit{functional object}. It first applies open-vocabulary segmentation to locate the contextual object, and then employs an MLLM to find the corresponding functional object. However, both approaches are designed specifically for the SceneFun3D scenario and fail to generalize effectively to other environments or datasets.

\section{Method}
\label{sec:method}
\paragraph{Problem Formulation}
Given a multi-view scene observation, composed of a set of $T$ RGB frames $\{I_t\}_{t=1}^T$, their corresponding depth maps $\{D_t\}_{t=1}^T$, camera poses $\{E_t\}_{t=1}^T$, camera intrinsics $K$, and a pre-reconstructed 3D point cloud $P$, our goal is to identify and segment a set of 3D instances $\mathcal{O} = \{o_1, o_2, ..., o_N\} \subset P$ that satisfy a flexible user task description $\mathcal{L}$. This task $\mathcal{L}$ can range from simple open-set category labels (e.g., "chair, table, etc.") to a complex, arbitrary natural language query (e.g., "Adjust the room's temperature using the radiator dial next to the bed").

\paragraph{Overview}
We present a unified framework that performs open-vocabulary 3D instance segmentation. Bootstrapped by sparse occupancy and appearance evidence, coarse 3D instances are associated across views through temporal consistency in 3D space; no hand-crafted heuristics are required.  
The resulting proposals are refined via 2D-3D consistency checks and an optional geometric enhancement step, while duplicate candidates are merged to remove redundancy.  
Finally, a multi-modal large language model classifies each instance, delivering accurate open-vocabulary recognition. An overview of our \textit{Proposal Generation} and \textit{Proposal Refinement} stages is shown in Fig.~\ref{fig:ppl}.

\subsection{Proposal Generation}
We introduce a unified design for 3D instance proposal mining from video, consisting of \textbf{2D Instance Initialization} and \textbf{Visual-Spatial Tracking}. First, \textbf{2D Instance Initialization} extracts occupancy and appearance cues from sparse frames and lifts them into coarse 3D instances. Then, \textbf{Visual-Spatial Tracking} exploits temporal consistency in 3D occupancy and appearance to associate instances across views, removing the need for hand-crafted heuristics. This streamlined pipeline yields robust instance-level representations without relying on explicit 3D priors.

\subsubsection{2D Instance Initialization}
We first harvest 2D instance cues that later guide 3D association.
The input video is temporally subsampled to a sparse frame set that still spans the scene while suppressing redundancy. 
Instance-level cues are gathered in two streams: occupancy and appearance. 
For occupancy, an open-vocabulary detector proposes, for every frame $I_i$, a set of bounding boxes $\{b_{i,j}\}$. 
Each box is forwarded to SAM to yield a foreground mask $m_{i,j}$. 
Using calibrated depth $D_i$, intrinsics $K$, and camera pose $E_i$, each box–mask pair is lifted into a coarse 3D point cloud. We then apply DBSCAN~\cite{ester1996density} to remove noise and recover a clean object-level point cloud $\mathcal{I}_{i,j}$.
For appearance, we compute mask-averaged DINO~\cite{oquab2023dinov2} features as the descriptor ${\mathbf{f}}_{i,j}$.

\subsubsection{Visual-Spatial Tracking}
Building upon conventional 2D multi-object tracking frameworks, our \textbf{Visual-Spatial Tracking} scheme aggregates multi-view information into a unified instance-level representation. 
Our method jointly exploits temporal consistency by leveraging coherence in both \textbf{3D spatial occupancy} and \textbf{visual appearance}.
By leveraging the natural temporal coherence across frames, the proposed tracking module removes the need for hand-crafted matching rules or label-assignment heuristics. 

Each object's 3D occupancy (the \textbf{spatial} cue) is encoded as a coarse voxel grid, and their geometric consistency is measured via an efficient voxel-level IoU metric. 
Meanwhile, appearance similarity (the \textbf{visual} cue) is evaluated using the cosine distance between instance-level feature embeddings. 
Combining these two cues, the pairwise similarity score between a new instance $\mathcal{I}_{i,*}$ and an active track $\mathcal{T}_{(i-1)}$ is formulated as
\begin{equation}
s = \alpha \cos(\mathbf{f}_T, \mathbf{f}_I) + (1 - \alpha) \mathrm{IoU}^{\mathrm{voxel}}(\mathcal{I}_{i,*}, \mathcal{T}_{(i-1)}),
\end{equation}
where $\alpha$ balances the influence of appearance and occupancy similarity, and $\mathbf{f}_T$, $\mathbf{f}_I$ denote the feature embeddings of $\mathcal{T}_{(i-1)}$ and $\mathcal{I}_{i,*}$, respectively. As illustrated in Fig.~\ref{fig:ppl}, even when the reconstructed chair point cloud of a new frame yields a low voxel IoU, the strong visual similarity captured by DINO features still enables correct association; conversely, two spatially separated backpacks may share high visual similarity, yet the IoU term effectively disambiguates them by enforcing geometric coherence.

During online tracking, each tracklet’s voxel representation is incrementally updated by merging the voxels of its newly matched detection, while its DINOv2-based descriptor is refreshed via an Exponential Moving Average. 
The bipartite association between instances and existing tracks is then solved using the Hungarian algorithm. 
After association, a match is considered successful if its similarity score exceeds a threshold $\tau_{\mathrm{match}}$; otherwise, the unmatched instance initializes a new tracklet.

\subsection{Proposal Refinement}
After Visual-Spatial Tracking, we obtain the coarse 3D proposals $\mathcal{P}_k$ for each tracked instance $k$. These proposals are initially formed by aggregating the frame-wise point clouds and matching them to the global scene point cloud $\mathcal{P}$ via $K$-nearest neighbors (KNN) association. However, due to the inherent mismatch between the temporally lifted frame-wise instances and the static scene representation, coupled with potential registration errors and accumulated depth noise, the resulting proposal $\mathcal{P}_k$ often contains significant noise and outliers. Therefore, a meticulously designed refinement stage is crucial to achieving accurate and clean 3D instance boundaries.

\subsubsection{Consistency Refinement}
Due to the inherent mismatch during frame-to-scene aggregation, $\mathcal{P}_k$ often suffers from partially missed or thin structures. To address this, we first optionally perform a point cloud expansion on $\mathcal{P}_k$: we incorporate neighboring points from the global scene $\mathcal{P}$ within a distance $\tau_{\mathrm{exp}}$ to form an expanded candidate set $\mathcal{P}'_k$.

Then, we propose Consistency Refinement which is designed to eliminate noisy points from this expanded set $\mathcal{P}'_k$ by leveraging a  2D-3D consistency metric, derived from projection and multi-view voting. Specifically, for a given object's expanded point cloud $\mathcal{P}'_k$, we project it onto 2D image planes and evaluate the 2D-3D consistency score $s_j$ of each 3D point $p_j \in \mathcal{P}'_k$ by verifying whether the projected coordinate lies within the corresponding 2D instance mask of each visible view (i.e., views where the point projects onto the image canvas) and computing the average voting score as follows:
\begin{equation}
    s_j = \frac{1}{|I^{p_j}_{vis}|} \sum_{i \in I^{p_j}_{vis}} \mathbb{I}\big[ \pi_i(p_j) \in m_{i,k} \big]
\end{equation}
where $I^{j}_{vis}$ denotes the set of visible frame indices for point $p_j$. The function $\pi_i(\cdot)$ represents the projection from the world coordinate system to the 2D image plane of the $i$-th frame, and $\mathbb{I}(\cdot)$ is the indicator function. Ultimately, we filter out noisy points with consistency scores below the threshold $\tau_{\mathrm{vis}}$ and retain the remaining ones, $\mathcal{P}^{\text{cr}}_{k} = \{ \mathbf{p}_j \in \mathcal{P}'_{k} \mid s_j \geq \tau_{\mathrm{vis}} \}$, as the denoised and structurally complete object point cloud.

\begin{table*}[t]
\centering

\begin{tabular}{c c| c c c| c c c } 
 \hline
 Method & Supervised 3D Mask  & \multicolumn{3}{c|}{\textbf{ScanNet200}} & \multicolumn{3}{c}{\textbf{Replica}} \\
 \hline
  & & AP  & AP\textsubscript{50} & AP\textsubscript{25} & AP  & AP\textsubscript{50} & AP\textsubscript{25} \\
 \hline
 Mask3D~\cite{schult2022mask3d}(Closed-vocabulary
 )  & - & 26.9 & 36.2 & 41.4 & - & -  & -  \\
 \hline
OpenIns3D \cite{huang2024openins3d} & \checkmark & 8.8 & 10.3 & 14.4  & - & -  & -  \\ 
OpenScene \cite{peng2023openscene}& \checkmark & 11.7 & 15.2 & 17.8   & - & -  & -  \\ 
OpenMask3D \cite{takmaz2023openmask3d} & \checkmark & 15.4 & 19.9 & 23.1  & 13.1 & 18.4 & 24.2 \\ 
Open3DIS* \cite{nguyen2024open3dis} & \checkmark & 23.7 & 29.4 & 32.8  & 18.4 & 23.8 & 28.2 \\ 
Any3DIS* \cite{nguyen2025any3dis}& \checkmark & 25.8 & - & -   & - & -  & - \\ 
OpenYOLO3D \cite{boudjoghra2024open} & \checkmark & 21.9 & 28.3 & 31.7 & 23.7 & 28.6 & 34.8 \\
Detail Matters \cite{jung2025details} & \checkmark & 25.8 & 32.5 & 36.2  & 22.6 & 31.7 & 37.7 \\
\hline
SAM3D \cite{guo2024sam}& $\times$ & 9.8 & 15.2 & 20.7  & - & - &  - \\ 
OVIR-3D \cite{lu2023ovir} & $\times$ & 9.3 & 18.7& 25.0  &11.0 & 20.5 & 27.5 \\ 
MaskClustering \cite{yan2024maskclustering} &$\times$ & 12.0 & 23.3 & 30.1 & - & - & - \\ 
SAI3D \cite{yin2024sai3d} & $\times$& 12.7 & 18.8 & 24.1 & - & - & - \\ 
SAM2Objects \cite{zhao2025sam2object} & $\times$ & 13.3 & 19.0 & 23.8 & - & - &  -\\
Open3DIS* \cite{nguyen2024open3dis} & $\times$ & 18.2 & 26.1 & 31.4  & 18.2  & 25.9 & 31.0  \\ 
Any3DIS* \cite{nguyen2025any3dis} & $\times$ & 19.1 & - & -  & - & - & -  \\ 
Detail Matters\cite{jung2025details} & $\times$  & 21.5 & 31.2 & 37.7 & 20.8 & 32.4 & 38.5 \\
\textbf{Ours}  & $\times$ & \textbf{26.0} & \textbf{37.7} & \textbf{45.4}  & \textbf{23.9}& \textbf{36.4} & \textbf{47.6} \\
 \hline
\end{tabular}

\caption{Results on \textbf{ScanNet200} validation set and Replica dataset. Bold entries indicate best performance. Compared with methods that do not rely on supervised 3D masks, our approach achieves superior results, especially in terms of $\text{AP}_{50}$ and $\text{AP}_{25}$.}
\label{table:scannet200_replica_results}
\end{table*}

\subsubsection{Optional Geometry Refinement}

After Consistency Refinement, boundaries may still look ragged or extend onto large planar regions due to mask leakage and depth noise. To enhance geometric coherence further, we introduce an optional Geometry Refinement that leverages superpoints~\cite{felzenszwalb2004efficient}. 

In general, we utilize a set of instance-associated superpoints for the instance re-composition. More specifically, for each superpoint $\mathcal{S}_j$, if the proportion of its constituent points contained in $\mathcal{P}_k^{cr}$ exceeds the threshold $\gamma$, the superpoint $\mathcal{S}_j$ is incorporated into the reconstructed instance $\mathcal{P}_k^{gr}$, as formalized in Equation~\ref{geo_ref}. 

\begin{equation}
\label{geo_ref}
    \mathcal{P}^{gr}_k = \{ \mathcal{S}_j \in \mathcal{S} \mid \frac{|\mathcal{S}_j \cap \mathcal{P}^{cr}_k|}{|\mathcal{S}_j|} \geq \gamma \}
\end{equation}

\subsubsection{Proposal Merging}

Tracking and lifting can produce duplicate or highly overlapping hypotheses for the same object (e.g., occlusions or short-track fragmentation). We resolve redundancies by merging instances based on voxel-point-set overlap, following a standard non-maximum suppression (NMS) pipeline. Each track’s length serves as its objectness score, and tracks are sorted accordingly. For each track, we compute the voxel-point-set IoU with previously retained objects. If the IoU exceeds a threshold $\tau_{\mathrm{merge}}$, the track is considered redundant—likely from fragmented observations of the same object—and its points are merged into the existing object.

\begin{figure}[ht]
\centering
\includegraphics[width=0.9\linewidth]{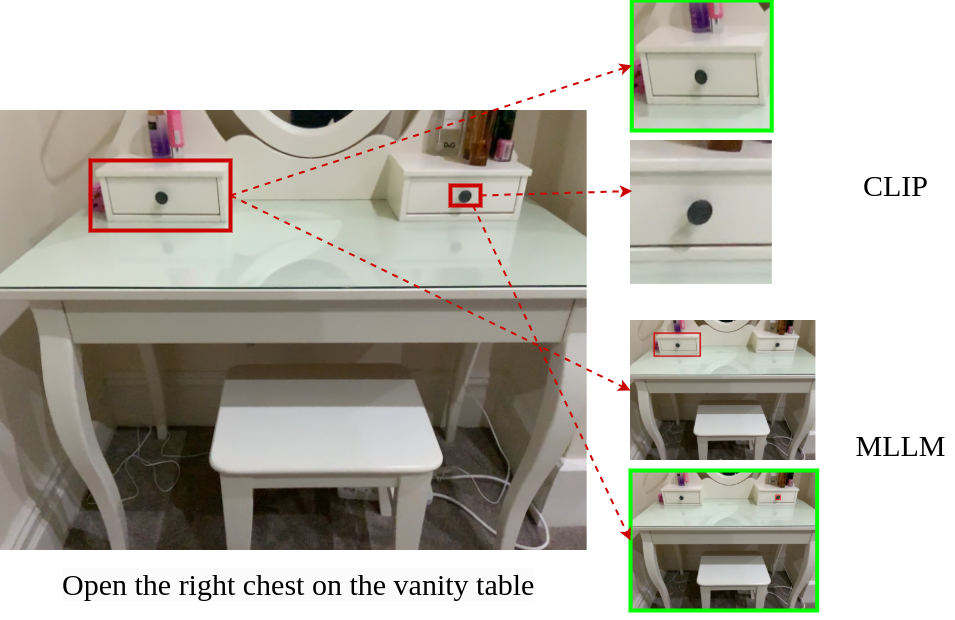} 
\caption{
Comparison of CLIP and MLLM on fine-grained, complex textual queries. CLIP receives cropped object regions, while MLLM processes the full image with the target highlighted by a red box. Green boxes indicate predicted instances for each method. MLLM better captures complex descriptions and produces more accurate predictions, whereas CLIP often fails.
}
\label{fig:clip_vs_vlm}
\end{figure}

\begin{figure*}[t]
\centering
\includegraphics[width=0.85\linewidth]{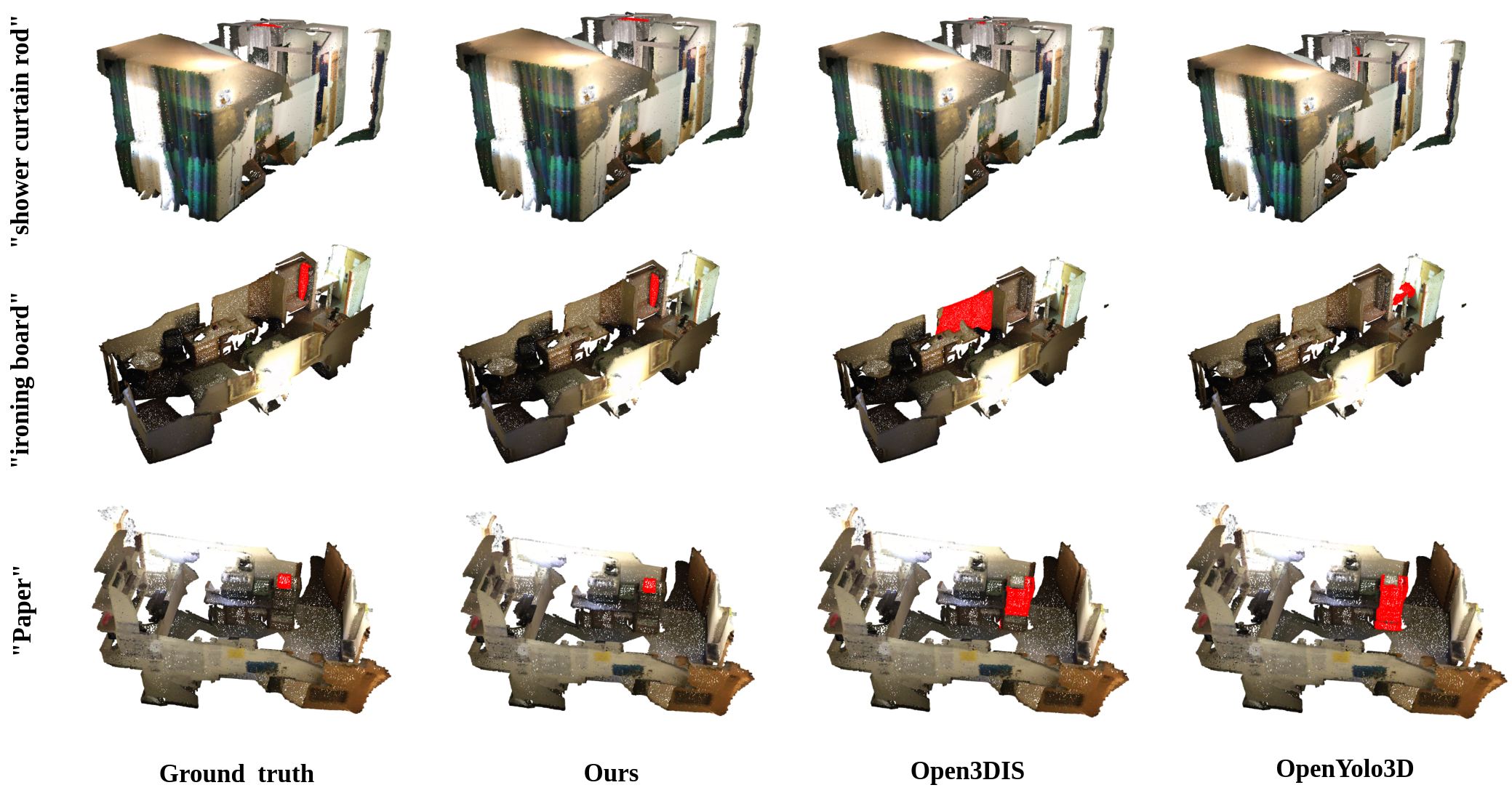} 
\caption{Comparative Visualization of Open-Vocabulary 3D Instance Segmentation on ScanNet200. Red masks denote query-correlated 3D instances. The comparative results demonstrate our method's clear advantage in addressing long-tail categories, showcasing improved segmentation accuracy and robustness when compared to baseline approaches.}
\label{fig:scannet200_cmp}
\end{figure*}

\subsection{Proposal Classification}
Previous open-vocabulary approaches typically use CLIP to associate textual queries with candidate proposals.
However, the limited pre-training objective and scale of CLIP constrain its ability to reason over fine-grained categories or complex textual descriptions. This limitation constitutes a major bottleneck in open-vocabulary perception.

To address this issue, we employ multi-modal large language model (MLLM) for the final classification stage.
To keep computation efficient, we leverage the tracklets established earlier and select a compact set of representative frames for each object. 

Specifically, we rank all camera views by the fraction of the object's reconstructed point cloud that falls within each camera frustum. Ties are broken using the projected mask area in the image, favoring views that observe the object frontally and at close range. As illustrated in \cref{fig:clip_vs_vlm}, we select the \text{top-}K views for each object and feed them into the MLLM. 
Each input consists of the full image where the target object is highlighted by a red bounding box. 
The MLLM is prompted to determine whether the highlighted instance belongs to any category in the restricted label set~$\mathcal{C}$, or semantically matches a user-provided textual description. 
By jointly reasoning over the top-$K$ representative images, the model achieves more robust and context-aware classification.

\section{Experiments}

\subsection{Experimental Setup}

\paragraph{Datasets}
We first evaluate our method on the widely used open-vocabulary 3D instance segmentation benchmarks ScanNet200~\cite{dai2017scannet} and Replica~\cite{replica19arxiv}, enabling fair comparisons with existing approaches. The ScanNet200 validation set consists of 312 scenes spanning 200 object categories, while Replica is a synthetic dataset comprising 8 scenes with 48 categories. Both datasets exhibit similar point cloud distributions and category distributions across scenes. Furthermore, to thoroughly assess the generalizability of our approach, we extend our evaluation to ScanNet++~\cite{ScanNet++} and SceneFun3D~\cite{delitzas2024scenefun3d}—two benchmarks that present substantially different and more challenging conditions. ScanNet++ provides increased complexity with 50 meticulously annotated scenes containing 1,554 fine-grained semantic labels. SceneFun3D challenges methods to perform language-guided segmentation of functional elements based on complex textual descriptions. It provides highly fine-grained annotations, including handles and buttons, and consists of 30 validation and 85 test scenes.

\paragraph{Metrics}
Our evaluation employs the standard average precision (AP) metrics: AP (computed across IoU thresholds from 0.5 to 0.95 with 0.05 increments), AP\textsubscript{50} (at an IoU threshold of 0.5), and AP\textsubscript{25} (at an IoU threshold of 0.25). Notably, we report results under the Top-1 evaluation protocol, as it aligns more closely with real-world application scenarios. Results obtained using the Top-k evaluation protocol—where each instance can have multiple predicted categories and the final result is derived from the top 300/600 predictions—are marked with an *.

\paragraph{Implementation Details} 
For the ScanNet200 and ScanNet++ datasets, we follow prior works and downsample the RGB-D frames by factors of 5 and 10, respectively.
We employ YOLO-World-X~\cite{cheng2024yolo} and SAM2-Large~\cite{ravi2024sam} as our open-vocabulary segmenter, and use DINOv2-Base as the backbone for semantic feature extraction. 
For the MLLM, we adopt Qwen3-VL-4B, which provides a favorable balance between performance and inference efficiency. For all benchmarks, we use the same hyperparameters for proposal generation, with $\tau_{\mathrm{match}}$ set to 0.4 and $\alpha$ set to 0.5.  
For proposal refinement, we default to  $\tau_{\mathrm{exp}} = 0.03~\mathrm{m}$, $\tau_{\mathrm{vis}} = 0.1$, $\tau_{\mathrm{merge}} = 0.6$ and $\gamma = 0.3$. 
For datasets where most objects are fully visible in the camera views, such as SceneFun3D, we adopt $\tau_{\mathrm{exp}} = 0$ and a higher threshold of $\tau_{\mathrm{vis}} = 0.8$. 
In addition, for the MLLM parameter $K$, we use $K=3$ on ScanNet200 and $K=1$ for all other datasets. 
More implementation details and ablation studies on hyperparameter sensitivity are provided in the supplementary material.

\begin{figure*}[ht]
\centering
\includegraphics[width=0.85\linewidth]{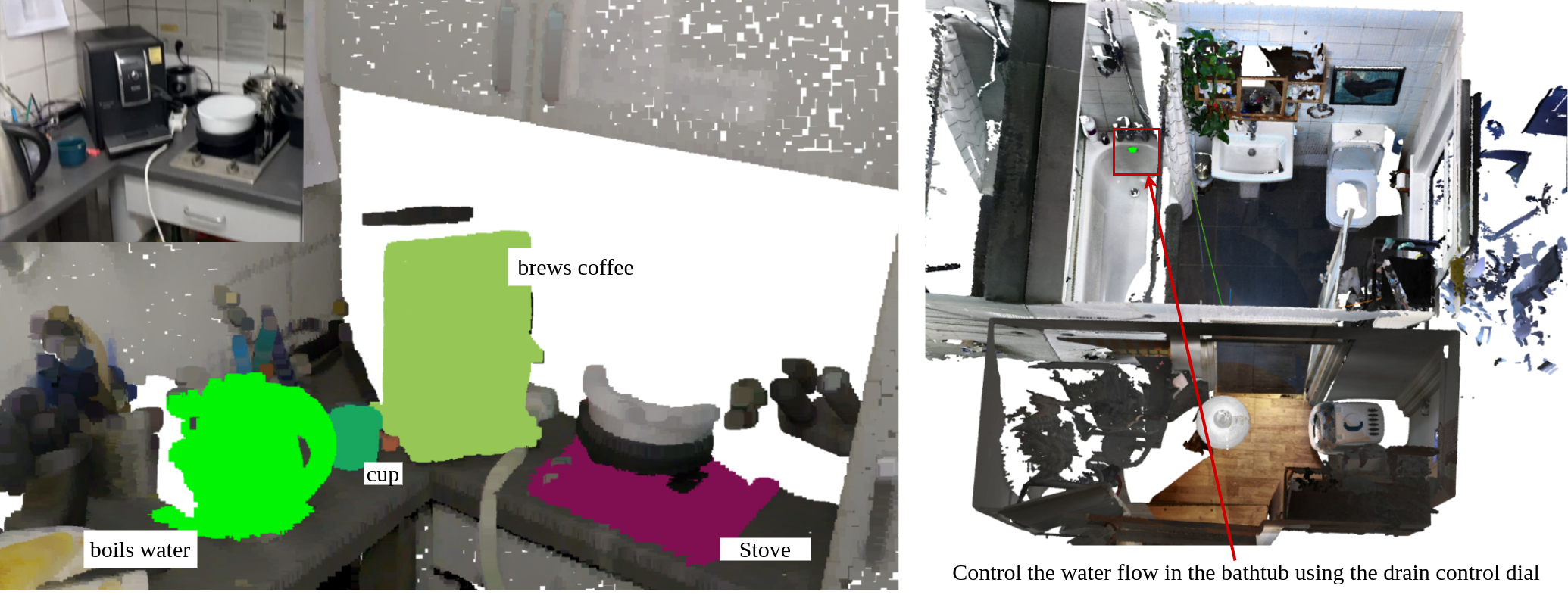} 
\caption{Qualitative Results on ScanNet++ (left) and SceneFun3D (right). We demonstrate instance segmentation results for various text queries. The visualizations demonstrate our method's capability to delineate objects with arbitrary geometries while maintaining recognition accuracy for specific textual descriptions.}
\label{fig:scannetpp_vis}
\end{figure*}

\subsection{Main Results}

\paragraph{Results on ScanNet200 and Replica} 
To ensure fair benchmarking against existing open-vocabulary 3D instance segmentation approaches, we perform comparative experiments on the ScanNet200 and  Replica datasets. Quantitative results are presented in Table~\ref{table:scannet200_replica_results}, and qualitative comparisons on ScanNet200 are shown in ~\cref{fig:scannet200_cmp}. Our method establishes a new state-of-the-art among open-vocabulary 3D instance segmentation approaches. When compared with methods that do not rely on supervised 3D masks from pre-trained models, our approach surpasses all previous SOTA results by a large margin, especially in terms of $\text{AP}_{50}$ and $\text{AP}_{25}$. On \textbf{ScanNet200}, we outperform the previous best by 6.5\% and 8.3\%, while on \textbf{Replica}, the improvements reach 4.0\% and 9.1\%, respectively. These results demonstrate that our generalizable framework consistently achieves significant advantages even on standard benchmarks.

\begin{table}[ht]
\centering
\begin{tabular}{c | c c c } 
 \hline
 Method & AP & AP\textsubscript{50} & AP\textsubscript{25}  \\
 \hline
 Segment3D~\cite{huang2024segment3d} & 10.1 & 17.7  & 20.2  \\
 Open3DIS~\cite{nguyen2024open3dis} & 11.9 & 18.1  & 21.7  \\
 Any3DIS~\cite{nguyen2025any3dis} & 12.9 & 19.0  & 21.9  \\
 \hline
\textbf{Ours} & \textbf{20.6} & \textbf{34.2}  & \textbf{43.4}  \\
 \hline
\end{tabular}

\caption{Results on the ScanNet++ dataset, evaluated on a subset of 100 classes.}
\vspace{-0.5cm}
\label{table:scannetpp_results}
\end{table}

\paragraph{Results on ScanNet++ and SceneFun3D}
To further validate the generalization of our method, we conduct additional experiments on the challenging ScanNet++ and SceneFun3D datasets. For ScanNet++, we follow Any3DIS~\cite{nguyen2025any3dis} by benchmarking a subset of 100 classes. Quantitative results are presented in Table~\ref{table:scannetpp_results} while Table~\ref{table:scenefun3d_results}, and qualitative comparisons are shown in ~\cref{fig:scannetpp_vis}. Our method achieves a substantial improvement of 8.3\% in AP over the previous state-of-the-art method Any3DIS on ScanNet++. On SceneFun3D, our approach outperforms OpenMask3D-F, which relies on supervised 3D proposals trained on SceneFun3D. Compared to Fun3DU, which also leverages MLLMs, our method achieves a 5.3\% improvement. Compared with ScanNet200 and Replica, both ScanNet++ and SceneFun3D present greater challenges due to their larger-scale point clouds and finer-grained object categories. For instance, the Mask3D network trained on ScanNet200 fails to generalize to ScanNet++, while SceneFun3D is a mesh-free dataset where superpoint-based proposal generation methods are inapplicable. These results collectively demonstrate the strong \textbf{accuracy} and \textbf{generalization} of our approach across diverse and challenging 3D benchmarks.

\begin{table}[t]
\centering
\begin{tabular}{c | c c } 
 \hline
 Method & AP\textsubscript{50} & AP\textsubscript{25}  \\
 \hline
 OpenMask3D~\cite{takmaz2023openmask3d} & 0.0  & 0.0  \\
 LERF~\cite{kerr2023lerf} & 4.9  & 11.3  \\
OpenMask3D-F~\cite{delitzas2024scenefun3d}& 8.0 & 17.5  \\
Fun3DU~\cite{corsetti2025functionality} & 3.6 & 8.8 \\
\hline
\textbf{Ours} & \textbf{8.9} & \textbf{18.5}  \\
 \hline
\end{tabular}
\caption{Results on the SceneFun3D test dataset. The reported Fun3DU results are obtained from the publicly available hidden test benchmark provided by SceneFun3D.}
\vspace{-0.2cm}
\label{table:scenefun3d_results}
\end{table}

\begin{table*}[t]
\centering
\begin{tabular}{c |c c c |c c c | c c c | c c } 
 \hline
method & \multicolumn{3}{c|}{ScanNet200} & \multicolumn{3}{c|}{Replica}  & \multicolumn{3}{c|}{ScanNet++}  & \multicolumn{2}{c}{SceneFun3D} \\
\hline
 & AP & AP\textsubscript{50} & AP\textsubscript{25}  & AP & AP\textsubscript{50} & AP\textsubscript{25} & AP & AP\textsubscript{50} & AP\textsubscript{25}  & AP\textsubscript{50} & AP\textsubscript{25}\\
 \hline
CLIP &  22.9 & 33.4 &  39.8 &  22.1 & 32.2 & 39.6 &  18.5 & 31.1 & 41.2 &  4.0 & 8.7 \\
MLLM &  \textbf{26.0} & \textbf{37.7}  &  \textbf{45.4} &  \textbf{23.9} & \textbf{36.4}& \textbf{47.6} & \textbf{20.6} & \textbf{34.2} & \textbf{43.4} & \textbf{9.8} & \textbf{20.6} \\
 \hline
\end{tabular}
\caption{Comparison between CLIP and multi-modal large language model (MLLM) classifiers on the validation sets of ScanNet200, Replica, ScanNet++, and SceneFun3D.}
\label{table:ablation_vlm}
\end{table*}

\subsection{Ablation Study}

\begin{table}[ht]
\centering
\begin{tabular}{c | c c c } 
 \hline
Method & AP & AP\textsubscript{50} & AP\textsubscript{25}  \\
 \hline
Baseline & \textbf{26.0} & \textbf{37.7} & \textbf{45.4} \\
w/o IoU score & 24.2 & 36.1 & 43.6 \\
w/o DINO score & 24.3 & 35.2 & 44.1 \\
w/o denoise & 25.5 & 37.6 & 45.2 \\
w/o mask-wise feature & 25.3 & 37.2 & 44.9 \\
 \hline
\end{tabular}
\caption{Ablation study on proposal generation evaluated on ScanNet200 validation set.}
\label{table:ablation_proposal_tracking}
\end{table}

\paragraph{Ablation Study on Proposal Generation.} 
We conduct ablation studies on the proposal generation stage to evaluate the contribution of different components. As shown in Table~\ref{table:ablation_proposal_tracking}, combining both the DINO score and the IoU score yields the best overall performance. Using only the visual (DINO) score tends to merge distinct objects with similar semantics (e.g., two spatially separated chairs) into a single track, reducing tracking precision. In contrast, relying solely on the IoU score requires setting an excessively high threshold to prevent merging adjacent instances, which in turn hampers cross-frame tracking.  
We further compare extracting DINO features from bounding boxes rather than instance masks. The results show that bbox-wise features lead to noticeable degradation, likely due to the inclusion of background noise and reduced robustness to occlusion. In addition, removing the denoising step on frame-wise point clouds also results in a clear performance drop, highlighting that cleaner per-frame proposals positively impact both tracking and segmentation accuracy.

\paragraph{Ablation Study on MLLM} 
To validate the effectiveness of employing a multi-modal large language model (MLLM) as the classifier, we compare it with the CLIP-based baseline used in prior works. Following OpenMask3D, we select the top-5 views per instance and extract multi-scale CLIP features, which are then averaged to form the final representation. As shown in Table~\ref{table:ablation_vlm}, replacing CLIP with an MLLM consistently improves performance across all datasets. The gain is particularly pronounced on SceneFun3D, where fine-grained textual understanding is required. In such scenarios, the MLLM demonstrates a superior capacity for semantic reasoning compared to CLIP.  
Moreover, even compared to the recent state-of-the-art method~\cite{jung2025details}, which employs Alpha-CLIP and aggregates top-20 or top-40 view features, our simple MLLM-based classifier achieves superior results, demonstrating both higher effectiveness and efficiency.

\paragraph{Ablation Study on Proposal Refinement} 
As shown in Table~\ref{table:ablation_proposal_refine}, our refinement module consistently improves performance across different datasets. In particular, for the SceneFun3D dataset, which mainly consists of small and fine-grained categories such as \textit{buttons} and \textit{handles}, even a single inaccurate segmentation within a track can significantly degrade performance, making the proposed Consistency Refinement module essential. For ScanNet200, the most notable improvements appear in $\text{AP}$ and $\text{AP}_{50}$, indicating that while our tracking framework achieves high recall, its precision remains limited without refinement. To further enhance precision, we incorporate superpoints—derived from mesh geometry—into the refinement stage. This addition helps produce more spatially coherent instance masks on datasets with mesh. Importantly, our framework remains compatible with mesh-free datasets such as SceneFun3D, where the superpoint-based refinement can be disabled without affecting the overall pipeline.  Finally, as shown in the results, proposal merging and filtering play a minor role for SceneFun3D. This is likely because most small objects remain visible across consecutive frames, and our tracking module rarely produces fragmented tracklets, making such merging unnecessary.

\begin{table}[t]
\centering
\begin{tabular}{c | c c c | c c} 
 \hline
 Method &\multicolumn{3}{c|}{ScanNet200}  & \multicolumn{2}{c}{SceneFun3D} \\
 \hline
 & AP & AP\textsubscript{50} & AP\textsubscript{25} & AP\textsubscript{50} & AP\textsubscript{25} \\
 \hline
Baseline &  13.4 & 28.5 & 41.3 & 2.0  & 7.0\\
 + CR &  15.0 & 30.3  & 42.8  &  9.5 & 20.1\\
 + GR & 24.9 & 35.2  & 41.6  & - & -\\
 + Merge & \textbf{26.0} & \textbf{37.7} & \textbf{45.4}  & \textbf{9.8} & \textbf{20.6} \\
 \hline
\end{tabular}
\caption{Ablation study on proposal refinement on ScanNet200 and SceneFun3D validation sets. 
CR: Consistency Refinement. GR: Geometry Refinement. 
For SceneFun3D, which does not provide mesh geometry, the GR module is omitted.
}
\label{table:ablation_proposal_refine}
\vspace{-0.3cm}
\end{table}

\section{Limitation and Conclusion}
Despite achieving state-of-the-art results, OpenTrack3D remains limited by its dependence on video quality and the noise introduced during 2D-to-3D lifting. Imperfect camera poses or depth maps can impair accurate 3D localization and cross-view consistency.

Looking ahead, we aim for two primary extensions. First, we will enhance the robustness of the lifting stage by integrating more advanced, unified 2D foundation models. Second, leveraging OpenTrack3D's demonstrated strong generalization capability, we envision applying our framework as a cost-effective tool for large-scale, open-vocabulary 3D data annotation. This line of research will ultimately accelerate the development of 3D-native open-vocabulary foundation models, bridging the current gap between generalized 2D and 3D vision.

\clearpage
\setcounter{page}{1}
\maketitlesupplementary

\section{More Implementation Details}
\label{sec:more_imp}

In this section, we provide additional implementation details of our MLLM components to improve understanding and reproducibility.
We deploy Qwen3-4B using vLLM~\cite{kwon2023efficient} on a single A100 GPU and use the default generation settings unless otherwise noted.

For the SceneFun3D~\cite{delitzas2024scenefun3d} dataset, the user-provided task descriptions cannot be directly processed by open-vocabulary detectors. To bridge this gap, we use the same MLLM to convert each textual query into a list of directly operable affordance nouns using the following prompt: \texttt{Extract directly operable affordance elements from the given task description (e.g., handle, knob, switch, latch, lever). Do not include whole objects. Output a comma-separated list of nouns. Task: \{\}.}
These extracted nouns are then fed into the open-vocabulary detector.

For datasets~\cite{dai2017scannet,replica19arxiv,ScanNet++} with predefined category sets, we adopt a unified prompt template:
\texttt{Identify the object shown in the red rectangle. Return the class name from \{\}. If no object is shown, return 'none'}
In contrast, for SceneFun3D, we employ a specialized prompt that maps all task descriptions in a scene to their corresponding task indices in a single forward pass:
\texttt{The interactive element is shown in the red rectangle. Determine whether the element in the image can accomplish any task in the task list. If so, return the task index; otherwise return 'no match'. Tasks: \{\}}.
This batch-style prompting significantly accelerates inference compared with querying each proposal independently.

\section{Hyperparameter Sensitivity Analysis}
\label{sec:add_results}

\begin{table}[th]
\centering
\begin{tabular}{c | c c c} 
\hline
$\tau_{\mathrm{match}}$ & AP & AP\textsubscript{50} & AP\textsubscript{25} \\
\hline
0.6 & 26.4 & 38.3 & 46.4 \\
0.5 & 28.2 & 39.6 & 47.9 \\
0.4 & \textbf{28.6} & \textbf{40.5} & \textbf{48.8} \\
0.3 & 27.3 & 38.2 & 45.9 \\
\hline
\end{tabular}
\caption{Sensitivity analysis of the $\tau_{\mathrm{match}}$ parameter on subsets of the ScanNet200 validation set.}

\label{table:hyper_match}
\end{table}

We analyze the sensitivity of key hyperparameters used in our framework. 
As shown in~\cref{table:hyper_match} and~\cref{table:hyper_refine}, the results remain stable across a wide range of values in both the Proposal Generation and Proposal Refinement modules.

We also evaluate the effect of varying the MLLM top-$K$ parameter, as summarized in~\cref{table:hyper_cls}. 
Larger $K$ generally improves performance but increases inference latency; in practice, users can choose an appropriate trade-off based on computational constraints.

\begin{table}[t]
\centering
\setlength{\tabcolsep}{4pt}
\begin{tabular}{c c | c c | c c | c c} 
\hline
$\tau_{\mathrm{exp}}$ & AP & $\tau_{\mathrm{vis}}$ & AP & $\gamma$ & AP & $\tau_{\mathrm{merge}}$ & AP \\
\hline
0.00 & 28.0 & 0.1 & 28.6 & 0.5 & 27.6 & 0.8 & 27.6 \\
0.03 & 28.2 & 0.2 & \textbf{29.1} & 0.4 & 28.6 & 0.7 & 28.4 \\
0.05 & \textbf{28.6} & 0.3 & 27.2 & 0.3 & \textbf{28.6} & 0.6 & \textbf{28.6} \\
0.07 & 28.0 & 0.4 & 24.9 & 0.2 & 27.0 & 0.5 & 27.0 \\
\hline
\end{tabular}
\caption{Sensitivity analysis of hyperparameters in the Proposal Refinement module on subsets of the ScanNet200 validation set.}

\label{table:hyper_refine}
\end{table}

\begin{table}[t]
\centering
\begin{tabular}{c | c c c | c c} 
\hline
& \multicolumn{3}{c|}{ScanNet200} & \multicolumn{2}{c}{SceneFun3D} \\
\hline
K & AP & AP\textsubscript{50} & AP\textsubscript{25} & AP\textsubscript{50} & AP\textsubscript{25} \\
\hline
1 & 28.6 & 40.5 & 48.8 & 9.8  & 20.6 \\
2 & 28.8 & 40.8 & 49.0 & 10.2 & 20.8 \\
3 & \textbf{28.9} & \textbf{41.8} & \textbf{49.5} & \textbf{10.5} & \textbf{21.2} \\
\hline
\end{tabular}
\caption{Ablation study on the MLLM top-$K$ parameter on subsets of the ScanNet200 and SceneFun3D validation sets.}
\label{table:hyper_cls}
\end{table}

\section{Runtime Analysis}
\label{sec:runtime}

We analyze the runtime of our framework and compare it with the previous state-of-the-art, DetailMatters~\cite{jung2025details}. 
Most of our runtime is spent on model inference. 
The cost of the 2D open-vocabulary segmenter and DINO scales with the number of video frames, while the proposal-classification cost (via either MLLM or CLIP~\cite{radford2021learning}) mainly depends on the number of proposals.

DetailMatters~\cite{jung2025details} follows the Open3DIS~\cite{nguyen2024open3dis} pipeline and employs Grounding-DINO~\cite{liu2024groundingdinomarryingdino} as its 2D open-vocabulary detector. Because our method additionally incorporates a DINO~\cite{oquab2023dinov2} model, we choose to pair it with a more lightweight 2D open-vocabulary detector, YOLO-World~\cite{cheng2024yolo}, to balance the runtime overhead. As shown in~\cref{table:time_ov}, despite using one additional model, our overall runtime is approximately half that of DetailMatters.

The gap becomes even more pronounced in the proposal-classification stage. 
DetailMatters extracts the top 20 images for each proposal and applies multi-scale cropping, resulting in roughly 60 Alpha-CLIP~\cite{sun2024alpha} forward passes per proposal, in addition to substantial preprocessing overhead. 
In contrast, our pipeline directly overlays a red bounding box on the original image to highlight the target and only feeds the top 3 images into the MLLM. As shown in~\cref{table:time_cls}, our approach achieves substantially lower inference latency compared to DetailMatters.

For fairness, all reported DetailMatters timings are obtained from our custom re-implementation, and are generally lower than the values reported in the original paper due to differences in implementation and hardware.

\begin{table}[t]
\centering
\setlength{\tabcolsep}{3pt}
\begin{tabular}{c | c c | c c} 
\hline
 & 2D OV-SEG & DINO & Frames & Total \\
\hline
DetailMatters~\cite{jung2025details} & 1.4 & - & 346 & 484 \\
Ours & 0.52 & 0.25 & 346 & 266 \\
\hline
\end{tabular}
\caption{Runtime of the 2D open-vocabulary segmenter and DINO. We report the per-frame runtime (in seconds), the average number of frames per scene, and the average runtime per scene on the ScanNet200 validation set.}

\label{table:time_ov}
\end{table}

\begin{table}[t]
\centering
\setlength{\tabcolsep}{4pt}
\begin{tabular}{c | c | c c} 
\hline
 & Classification & Proposals & Total \\
\hline
DetailMatters~\cite{jung2025details} & 5.9 & 56 & 330.4 \\
Ours ($K{=}3$) & 1.6 & 56 & 89.6 \\
Ours ($K{=}1$) & 1.1 & 56 & 61.6 \\
\hline
\end{tabular}
\caption{Runtime of the proposal-classification stage. We report the per-proposal runtime (in seconds), the average number of proposals per scene, and the average runtime per scene on the ScanNet200 validation set.}

\label{table:time_cls}
\end{table}


{
    \small
    \bibliographystyle{ieeenat_fullname}
    \bibliography{main}
}


\end{document}